# Anomaly Detection Using the Dempster-Shafer Method

Qi Chen and Uwe Aickelin

*Abstract*—In this paper, we implement an anomaly detection system using the Dempster-Shafer method. Using two standard benchmark problems we show that by combining multiple signals it is possible to achieve better results than by using a single signal. We further show that by applying this approach to a real-world email dataset the algorithm works for email worm detection. Dempster-Shafer can be a promising method for anomaly detection problems with multiple features (data sources), and two or more classes.

## I. INTRODUCTION

Intrusion Detection Systems (IDSs) play a pivotal role within network security [1]. IDSs are one of many tools used to detect attacks and intruders of computer systems. It is important to note that the purpose of IDSs is not to prevent the entry of intruders to a system, but to notify the administrator of any observed intruders.

IDS techniques can be categorised as either misuse detectors or anomaly detectors. Misuse detection systems, such as Snort [2], rely on intrusion signatures to detect an attack. Such signatures are stored in a database, which relies on frequent updates to remain functional. System behaviours are matched against the signatures within the database. If a successful match is formed, an alert is generated. An administrator can use these alerts to investigate the potential problem, and generate appropriate responses. However, like many anti-virus scanners, misuse-detectors rely on continual updates of the signature database. Hence the main drawback with this paradigm is that it will never detect 'day-zero' intrusions to which signatures have not yet been created.

Conversely, anomaly detection techniques generate profiles of normal behaviour. Deviations from the 'normal profile' result in the generation of alerts, which are used by the system administrator for audit purposes. The major advantage of anomaly detection systems is that novel attacks can be detected. Unfortunately, the profiles are not always accurate, as user behaviour changes over time. This can lead to the generation of false positive alerts, when previously unseen user behaviour occurs for legitimate reasons. The false positive rate can be sufficiently high that the anomaly detection system can be flooded by these alerts, forcing the administrator to either ignore the alerts or disable the system. Our work is part of the research to reduce the number of false alerts produced by anomaly detection systems.

Q. Chen is with the school of Computer Science & IT, University Of Nottingham, Nottingham, NG8 1BB, UK. (phone: 0115-951-4247; fax: 0115-951-4254; e-mail: qxc@ cs.nott.ac.uk).

U. Aickelin is with the school of Computer Science & IT, University Of Nottingham, Nottingham, NG8 1BB, UK. (e-mail: uxa@cs.nott.ac.uk).

A considerable number of anomaly detection systems have been developed. Examples include [3] who employed statistical and grammatical metrics to detect anomalies within system calls, and [4] which used an immune-inspired system to detect abnormal processes.

Anomaly detection is not restricted to computer security. Other applications such as threat assessment and medical diagnosis rely on detecting deviations within dynamic environments. One technique used for detection is *Multisensor Data Fusion* [5]. This is a form of signal processing, where data from multiple sources is used for analysis. The Dempster-Shafer theory of inference is a statistical method, considered as a generalised Bayesian theory, which can be used to combine multiple streams of input data. We believe that the Dempster-Shafer method can be successfully applied to anomaly detection through assigning 'belief values' to inputs from various data sources.

The remainder of this paper is organised as follows. Section II discusses the fundamentals of the Dempster-Shafer Theory and its advantages and disadvantages. An anomaly detection approach using the Dempster-Shafer theory is presented in III. We give some experimental results for two standard benchmark problems in IV and V. These two datasets are the Wisconsin Breast Cancer Dataset and the Iris dataset of the UCI Machine Learning Repository [6]. The experiment results for the email worm dataset (collected by our colleague) are described in VI. VII concludes the paper.

## II. THE DEMPSTER-SHAFER (D-S) THEORY

The Dempster-Shafer (D-S) theory is a mathematical theory of evidence, introduced in the 1960's by Arthur Dempster [7] and developed in the 1970's by Glenn Shafer [8]. The D-S Theory is viewed as a mechanism for reasoning under epistemic (knowledge) uncertainty. The part of the D-S theory which is of direct relevance to our work is *the Dempster's rule of combination*. We present some essential mathematical terminologies in section A, before we introduce *the Dempster's rule of combination* in B. We introduce the advantages and disadvantages of D-S in C.

### A. Basic mathematical terminology

**Frame of discernment ($\Theta$)** is a finite set mutually exclusive propositions and hypotheses about some problem domain.
**Basic probability assignment (bpa)** is stated in [8] as : "If $\Theta$ is a frame of discernment, then a function m: $2^{\Theta} \rightarrow [0,1]$ is called a *basic probability assignment* whenever

$$m(\phi) = 0 \qquad (1)$$

and

$$\sum_{A \subset \Theta} m(A) = 1. \qquad (2)$$

The mass value of A (m(A)) is also called A's *basic probability number*, and it is understood to be the measure of the belief that is committed exactly to A."

**Belief function (Bel)** is a belief measure of a proposition A, and it sums the mass value of all the non-empty subsets of A. This subset is also called the focal element of the *Bel*.

$$Bel(A) = \sum_{B \subseteq A} m(B) \qquad (3)$$

**Plausibility function (Pl)** takes into account all the elements related to A (either supported by evidence or unknown).

$$Pl(A) = 1 - Bel(\neg A) \qquad (4)$$

For the subset A, Bel(A) and Pl(A) represent upper and lower belief bounds, and the interval [Bel(A), Pl(A)] represents the belief range. The relationships between Bel value, Pl value and uncertainty are described in Figure 1.

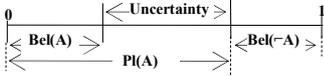

Fig. 1. The uncertainty interval for a hypothesis [9]

### B. The Dempster's Rule of Combination

$$m_{12}(A) = \frac{\sum_{B \cap C = A} m_1(B) m_2(C)}{1 - \sum_{B \cap C = \phi} m_1(B) m_2(C)} \qquad (5)$$

We can use Dempster's rule of combination to combine the mass values of all features from each individual sensor to achieve the overall summary mass values for each sensor. These summary values from all sensors are combined to give the summary mass values for the system.

Initially, the *bpa*s are used to assign the mass values to appropriate hypothesis. Then the resulting mass values are used to calculate the belief for the appropriate hypothesis. Finally all beliefs are combined with Dempster's rule of combination to gain the overview belief for the appropriate hypothesis, as shown in Equation (5).

### C. Advantages and Disadvantages of D-S

The main advantage of D-S is that no *a priori* knowledge is required, making it potentially suitable for anomaly detection of previously unseen information. Another advantage is that a value for ignorance can be expressed, giving information on the uncertainty of a situation. Bayesian inference requires a priori knowledge and does not allow allocating probability to ignorance. It can only express the probability of an event being either abnormal or normal. It is our opinion that a Bayesian approach is not always suitable for anomaly detection because pre-existing knowledge may not always be provided. In particular, if the aim is to detect previously unseen attacks, then a system which relies on existing knowledge cannot be used.

There are two major problems associated with D-S: the computation complexity and conflicting beliefs management. The computational complexity increases exponentially with the number of *frames of discernment ($\Theta$)*. If there are $n$ elements in $\Theta$, there will be up to $2^{n-1}$ focal elements for the mass function. The combination of two mass functions needs the computation of up to $2^n$ intersections. To overcome this, various algorithms, such as [10] and [11], have been suggested to reduce the focal element number in the involved mass functions. For anomaly detection, the resulting computation complexity is low, as the *frame of discernment* consists of only two elements (normal and abnormal). There are up to three focal elements of belief functions: {normal}, {abnormal}, and {normal or abnormal} (i.e. the uncertainty), resulting in low computation complexity

The Dempster's rule of combination redistributes the mass values of empty propositions to non-empty propositions, also known as normalization step, due to the definition of the mass function. This sometimes leads to erroneous results, which causes the conflicting management problem. In order to solve this problem, some alternative combination rules have been proposed, as in [12] and [13], but none have yet been accepted as a standard method. In order to illustrate this problem, consider the following example: a car window has been broken, and the culprit needs to be identified. There are three suspicious people (Jon, Mary, and Mike) and two witnesses (Witness1 and Witness2). Witness1 assigns "Jon broke it" with a mass value of 0.9, and "Mary broke it" with a mass value of 0.1; witness2 assigns "Mike broke it" with a mass value of 0.9, and "Mary broke it" with a mass value of 0.1. Both witnesses assign a very small mass *value* to "Mary broke it". Applying the Dempster's rule of combination for "Mary broke the window", returns a value of 1, which is not accurate. This is because the mass *value* can be affected by taking into account conflicting opinions of multiple sources. For our anomaly detection application, each *bpa* will assign a non-zero mass value to {normal or abnormal} as the error rate; therefore we will not face any belief conflict problems.

In summary, the D-S method is a combination of a theory of evidence and probable reasoning, to derive a belief that an event has occurred. Individual beliefs are updated and combined to give a belief of an event occurring in the system as a whole. Though a hotly debated point, D-S has advantages over Bayesian techniques when applied to anomaly detection as described above. For our application, each *bpa* will assign a non-zero mass value to {normal or abnormal}, this avoids any belief conflict problems.

### III. THE APPLICATION OF D-S IN ANOMALY DETECTION

We implemented a D-S system and applied it to two standard benchmark problems of the UCI datasets [6], the Wisconsin Breast Cancer Dataset (*WBCD*) and the Iris Dataset, and one email dataset made by our colleague. Two standard benchmark dataset are chosen to compare our approach with the performance of other algorithms, and to investigate whether it is possible to achieve good results by combining various features using D-S. The email dataset is chosen, because it is in our interested application area.

The anomaly detection system uses a training process to derive thresholds from the training data, and detects an event as normal or abnormal (as shown in figure 2). The *bpa* functions are built based on these thresholds for the purpose of assigning mass values. The anomaly detection approach is demonstrated in figure 3. The data from various sources are processed and sent to corresponding *bpa* assignment functions. The mass values for each hypothesis are generated and sent to D-S combination component. This component uses the Dempster's rule of combination to combine all mass values; and generate the overall mass values for each hypothesis. The data can be detected as normal or abnormal based on the overall mass values for each hypothesis.

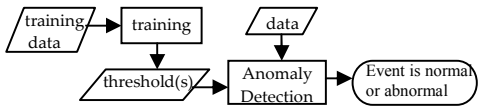

Fig. 2. Data flow of the Anomaly Detection System

All experiments for the three chosen datasets were executed on an Intel Pentium 4 CPU, 1.5G Hz, 256MB RAM, Windows 2000 platform computer. The system was coded using Java 2 platform, Standard Edition (J2SE) 1.4.0. The execution times (average running time of 10 runs) for the three datasets are: 30 seconds for the WBCD, 25 seconds for the Iris dataset, and 12 seconds for the email dataset.

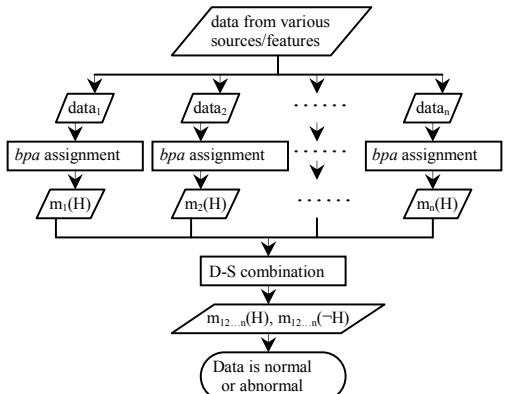

Fig. 3. Anomaly Detection Approach

## IV. EXPERIMENTS WITH THE WBCD

### A. The Wisconsin breast cancer dataset (WBCD)

The WBCD is a standard benchmark dataset of the UCI Machine Learning Repository [6]. This dataset is chosen for two objectives. One is to compare our approach with the performance of other algorithms. The other is to investigate whether it is possible to achieve good results by combining multiple features using D-S, without excessive manual intervention or domain knowledge based parameter tuning.

The WBCD contains 699 data items: 241 malignant items (abnormal data), and 458 benign items (normal data). This dataset has nine features; all features are normalised integers in the range between 1 and 10. A, B, C, D, E, F, G, H and I are used to represent the biological features of A: Clump Thickness, B: Uniformity of Cell Size, C: Uniformity of Cell Shape, D: Marginal Adhesion, E: Single Epithelial Cell Size, F: Bare Nuclei, G: Bland Chromatin, H: Normal Nucleoli, and I: Mitoses, respectively. There are 16 instances; each contains a single missing (i.e. unavailable) attribute value. Our D-S based anomaly detection system has the ability to cope with this problem by omitting, i.e. not combining, the missing values of the corresponding data items. This is an advantage of D-S over other approaches, such as [14] [15], which have to exclude the 16 items with missing values.

For the WBCD, the frame of discernment of the system is {normal, abnormal}. The *bpa* function and the *threshold* settings are illustrated in the next section.

### B. The classification approach

We use ten fold cross validation in our experiment. The dataset is divided into ten subsets of approximately equal size (one subset size is either 69 or 70). Each time we use the data of one subset as test data, and the data of the other nine subsets as training data. The training data is used to obtain the modified median threshold to build the *bpa* functions. The dataset size is 699, so the training data size is either 630 or 629. The proportional distribution of the WBCD is 65.5%:34.5% (normal: abnormal). We order the training data feature values from small to large based on each feature. If the training data size is 630, the $413^{th}$ small value of one feature is chosen as the modified median threshold. If the training data size is 629, the $412^{th}$ small value of one feature is chosen as the modified median threshold. We use a general assumption that the lower value items tend to be normal data.

Then the *bpa* function for each feature is :

$$\begin{cases} m(normal) = (1 + e^{(value - threshold)})^{-1} \\ m(abnormal) = 1 - m(normal) \end{cases} \quad (6).$$

Figure 4 shows a graphical illustration of the shapes of functions using a sample threshold of 5. Note that for the problems we study, all data items are integers and hence the functions consist of discrete values only.

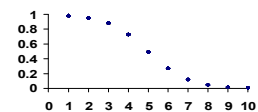

Fig. 4. Part of the *bpa* function for the WBCD, the x-axis shows feature values, y-axis shows mass values

All thresholds for nine features are found, and the *bpa* functions are built for each feature. For each data item, the *bpa* functions are used to assign the mass values for each feature based on that feature value. For that data item, all mass values are combined to obtain the overall mass values of the hypothesis normal and of the hypothesis abnormal. If the mass value of the 'abnormal' hypothesis is bigger than the mass value of the 'normal' hypothesis, then it is classified as abnormal; otherwise it is classified as normal.

### C. Experimental results for the WBCD

To judge the quality of results, we compare the

classification accuracies, based on the following definition:

classification accuracy = $\frac{\text{Number of correctly classified items}}{\text{Total number of items}}$.

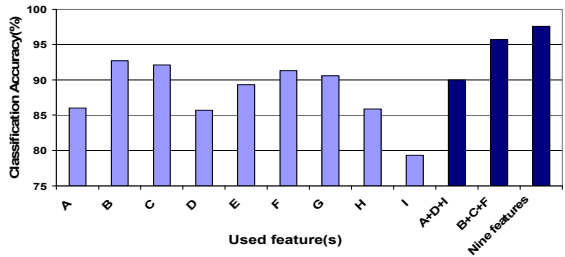

Fig. 5. Classification accuracies with various features for the WBCD

Figure 5 shows that feature A (classification rate 86.0%), D (85.7%) and I (79.3%) give the poorest performance when using only one feature at a time. The result when combing multiple features (A, D and I) together (90.0%) is better than using either A, D or I alone. Features B (92.7%), C (92.1%), and F (91.3%) are the best three features when using only one feature at a time. Similarly the result of combining these three (95.7%) is better than using either of them alone. The result of using all nine features (97.6%) is better than any other combination of features.

Our first hypothesis that combining features using D-S improves accuracy is proven correct for the WBCD Moreover, our second assumption is also proven correct, i.e. a few badly chosen features do not negatively influence the results, as long as most chosen features are suitable. These two characteristics make D-S very amenable for solving real-world IDS problems.

### D. Comparison with other methods

To appreciate the high quality of our results, we provide a comparison with other published results. [14] used a generalized rank nearest neighbour rule and achieved a classification rate of 96.17%; also [15] used a fuzzy classification method, with a best result of a classification rate equal to 96.7%. Both methods ignore the 16 WBCD data items with missing feature values. Our classification rate compares favourably with these being 97.6% (including all data with missing feature values). The ability to deal with missing values is important for network security problems. Our method has the advantage of having such ability.

## V. EXPERIMENTS WITH THE IRIS PLANT DATASET

### A. The Iris plant dataset

The Iris plant dataset is another standard benchmark problem of the UCI datasets [6]. This dataset is chosen because it has fewer features and more classes than the WBCD. This will confirm whether D-S can work on problems with fewer features and more classes.

This dataset has 150 instances with the following four numeric features: sepal length in cm; sepal width in cm; petal length in cm; and petal width in cm. The dataset also contains one predictable feature, namely the class label. These 150 instances are of three classes (plant type), *Iris Setosa, Iris Versicolour,* and *Iris Virginica*, with each class containing 50 instances.

### B. The classification approach

The Iris instances distribution overlapping information, based on individual feature, is used to roughly classify the Iris data (as shown in Figure 6). A number of items are not classified into a single class, such as either Setosa or Versicolour. For such data items, we use the difference between a data item value and the mean value of the selected suitable feature to provide classification into individual single classes. This classification approach is achieved in three steps, as described below.

In the first step, the system use *bpa A* to assign mass values to all the four features of one data item based on the boundary information. For this data item, the system combines the mass values using the Dempster's rule of combination, and then generates the overall mass values and belief values for all possible. There are seven possible hypotheses for the Iris dataset: {Setosa}; {Versicolour}; {Virginica}; {Setosa, Versicolour}; {Setosa, Virginica}; {Versicolour, Virginica}; and {Setosa, Versicolour, Virginica}. The data item is classified to the hypothesis with the highest belief value. If with the results of first step, the data item is not classified to a single class, such as {Setosa, Versicolour}, then the system uses the second step to classify it to a *single* class. In the second step, initially the most suitable feature is selected. Following this, the system uses the *bpa B* to assign mass values based on the distance to the mean values of the three classes of that feature. In the third step, the system combines the mass values of step one and step two. The overall mass values and the belief values are calculated. The items are classified to the hypothesis with the highest belief values.

We use ten fold cross validation in our experiment. The dataset is divided into ten subsets of equal size, with nine out of ten subsets comprising training data, with the remaining subset used as test data. The training data is used to obtain the thresholds to build the *bpa* assignment functions.

The following parts of section V are organised as below. Section B.1 details how to use *bpa A* to assign mass values based on the boundary information. In section B.3, we demonstrate the use of *bpa B* assigning mass values based on the differences between a feature value and the mean feature values of three classes. Finally the selection of suitable features is described in section B.2.

### B.1 bpa (basic probability assignment) function A

Firstly, we need to find the maximum and minimum values for each class based on one feature of the training data. Then we calculate the overlapping part for the three classes, to obtain the boundary information for each class, based on this feature of the training data, as shown in Figure 6. For Figure 6 and 7, each of the vertical lines is the value range for one class of one feature; the horizontal lines are used for comparison to calculate the overlap.

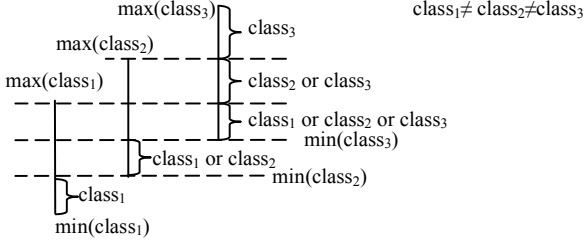

$class_1, class_2, class_3 \in \{\{Setosa\}, \{Versicolour\}, \{Viginica\}\}$.

Fig. 6. Example: for one feature, how to calculate the three class overlap

We use the example of Figure 6 to illustrate how to calculate the overlap. If the value is less than min(class$_2$), and greater than or equal to min(class$_1$), the data item must belong to class$_1$. This is because all the values of the data items belonging to class 2 should not be less than min(class$_2$). Similarly, the values of the data items belonging to class$_3$ should be not less than min(class$_3$). In this case, the min(class$_3$) is bigger than min(class$_2$), so the values of the data items belong to class 2 or class 3 should not be less than min(class$_2$). For the same reason, data items with values greater than max(class$_2$), belong to class$_3$. Data items with values between min(class$_2$) and min(class$_3$) belong to class 1 or class 2. Data items with value between min(class$_3$) and max(class$_1$) belong to class 1 or class 2 or class 3. For data items containing value between max(class$_1$) and max(class$_2$) belong to class 2 or class 3.

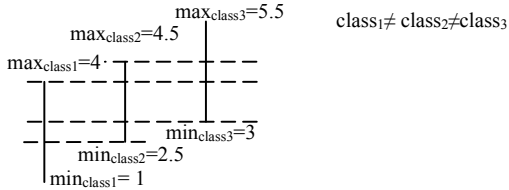

Fig. 7. Example of minimum, maximum value settings for Fig. 6.

In Figure 7, we set example maximum and minimum values, to illustrate the assignment of the mass values based on the boundary information. For one feature, if the feature value is less than min(class$_1$)=1, then that data item is classified as class$_1$. We assign m(class$_1$)=0.9, and m($\Theta$) = 0.1 based on that feature alone. As nothing is hundred percent accurate, we think the trustiness of this measurement is 0.9, and set the uncertainty ( m($\Theta$) ) as 0.1.

For each feature, we have the *bpa* function A:

$$\begin{cases} if\ value \in (-\infty, 2.5), m(class1) = 0.9, m(\Theta) = 0.1; \\ if\ value \in [2.5, 3), m(class1 \cup class2) = 0.9, m(\Theta) = 0.1; \\ if\ value \in [3, 4], m(\Theta) = 1; \\ if\ value \in (4, 4.5], m(class2 \cup class3) = 0.9, m(\Theta) = 0.1; \\ if\ value \in (4.5, +\infty), m(class3) = 0.9, m(\Theta) = 0.1. \end{cases}$$

In the first step, we apply *bpa* function A to each feature of one data item, and use the Dempster's rule of combination to combine the mass values of the four features. This generates the overall mass values for that data item. The overall classification is decided based on the overall mass values. If the data item is not classified to a single class, then we will use the second step.

*B.2 Feature selection*

Suitable features must be selected in order to separate two or three classes using the difference between the data item value and the mean value of the three classes. A feature is required with the following characteristics: the data feature values of one single class are close together; and the values of two classes viewed as a group are far apart. This is achieved by calculating the standard deviation for the two classes, and the standard deviation for the union of these two classes. This is defined as the Feature Selection Value (*FSV*), shown in Equation 7. The feature with the smallest *FSV* is chosen as the suitable feature.

The Feature Selection Value (FSV) for n (a natural number) classes is:

$$FSV = \frac{sd(class_1) \times sd(class_2) \times \cdots \times sd(class_n)}{sd(class_1 \cup class_2 \cup \cdots \cup class_n)} \quad (7)$$

For example, to separate the class Setosa and the class Versicolour, we select the feature with the smallest $FSV = \frac{sd(Setosa) \times sd(Versicolour)}{sd(Setosa \cup Versicolour)}$.

*B.3 bpa (basic probability assignment) function B*

In the second step, a suitable feature is selected and the *bpa* function B is used to assign mass values. We build *bpa* function B based on the information of the absolute distances (defined as difference in Equation 8) between the data item value of the chosen feature and the mean feature value of each class. Here, we want to classify the data item as the class with the smallest difference.

$$difference = |value - mean\ of\ one\ class| \quad (8).$$

The information of *bpa* B is viewed as less important than the information of *bpa* A. The mass value of one affected hypothesis is set as 0.8, the uncertainty as 0.2.

We have the following *bpa* function B:

$$\begin{cases} if\ Setosa\ has\ the\ smallest\ difference,\ then\ set \\ m(Setosa)=0.8,\ m(\Theta) = 0.2; \\ if\ Versicolour\ has\ the\ smallest\ difference,\ then\ set \\ m(Versicolour)=0.8,\ m(\Theta) = 0.2; \\ if\ Virginica\ has\ the\ smallest\ difference,\ then\ set \\ m(Virginica)=0.8,\ m(\Theta) = 0.2. \end{cases}$$

C. *Experimental results with the Iris Plant dataset*

The classification accuracy with the Iris plant dataset is 95.47% ± 0.48% (of ten runs). Table 1 shows three out of ten of the experimental results (chosen randomly for illustrative purpose). These results are based on the whole application approach with detailed error information and results. The table label meanings are explained below.

- 'Id': one item's identification number. 1-50: ids of Setosa, 51-100: ids of Versicolour, 101-150: ids of Virginica.
- 'Correct(1st)': the number of correctly classified items using the 1$^{st}$ step with the boundary information.

- 'Errors(1st)': the number of errors caused by the first step which only use the boundary information.
- 'In two(1st)': the number of date items, whose results are not in a single class after the first step.
- 'Errors(2nd)': the errors caused by the second step which use the "difference" information

TABLE 1 THE IRIS PLAN EXPERIMENTS USING THE WHOLE APPROACH

| 1st Run of the experiments—classification accuracy=96.6667 | | | | |
|---|---|---|---|---|
| Id | Correct(1st) | Errors(1st) | In two(1st) | Errors(2nd) |
| 1-50 | 50 | 0 | 0 | 0 |
| 51-100 | 35 | 2 (Id= 71,86) | 13 | Id=78 |
| 101-150 | 42 | 2(Id=107,120) | 6 | 0 |
| 2nd Run of the experiments—classification accuracy=95.3333% | | | | |
| Id | Correct(1st) | Errors(1st) | In two(1st) | Errors(2nd) |
| 1-50 | 50 | 0 | 0 | 0 |
| 51-100 | 33 | 4(Id=51,71,84, 86) | 13 | Id=78 |
| 101-150 | 42 | 2(Id=107,120) | 6 | 0 |
| 3rd Run of the experiments—classification accuracy=94.6667% | | | | |
| Id | Correct(1st) | Errors(1st) | In two(1st) | Errors(2nd) |
| 1-50 | 50 | 0 | 0 | 0 |
| 51-100 | 34 | 5(Id=51,57,71,84,86) | 11 | Id=78 |
| 101-150 | 43 | 2(Id=107,120) | 5 | 0 |

The data items 71, 86, 107 and 120 are wrongly classified in all these three runs, due to the first step of the classification approach. The 78 is wrongly classified in all three runs, due to the second step of the classification.

In Table 2, the parameters used in the 2nd run of the experiments for data 86 are shown. We use this example to show how some mistakes may occur. 86: F1-6 (classified as $class_{23}$); F2-3.4($class_{13}$); F3-4.5($class_{23}$); and F4-1.6 ($class_{23}$). It is classified as class 3 using the 1st step, which is wrong. F1, F3, and F4 express belief of $class_{23}$, and F2 expresses belief of $class_{13}$. The combination result is $class_3$. One plausible reason is that the training data does not include all the information of the test data; perhaps the parameters are not totally accurate. For the same reason, the data item 78 is wrongly classified due to the 2nd step. This can be improved by combining 1st and 2nd steps together. If we use 1st and 2nd step together for all the data, the expected results can be improved. We can also use a finer grained model to extract these individual features of the Iris data, which can also lead to better performance.

TABLE 2 THE PARAMETERS OF THE TRAINING SET FOR THE DATA ITEM 86

|  |  | $Class_1$ | $Class_2$ | $Class_3$ |
|---|---|---|---|---|
| Feature1 (F1) | max | 5.8 | 6.9 | 7.9 |
|  | min | 4.3 | 4.9 | 4.9 |
| Feature2 (F2) | max | 4.4 | 3.3 | 3.8 |
|  | min | 2.3 | 2.0 | 2.2 |
| Feature3 (F3) | max | 1.9 | 5.1 | 6.7 |
|  | min | 1.0 | 3.3 | 4.5 |
| Feature4 (F4) | max | 0.6 | 1.7 | 2.5 |
|  | min | 0.1 | 1 | 1.4 |

### D. Comparison with other methods

The classification accuracy of our method applied to the Iris data is 95.47%±0.48% over 10 runs. It is similar as published results of other established methods (whose results are between 94.67% and 97.33%) [16]. This demonstrates the ability of D-S to successfully classify items within datasets comprising few features and multiple classes.

## VI. EXPERIMENTS WITH EMAIL DATASETS

We have confirmed the potential of D-S to produce robust, high quality results. Then we turn our attention to the problem of worm detection. Due to the lack of datasets, we have derived our own data suitable for the detection of email worms. Worm detection forms a large subset of computer security and provides us with a managed problem to solve.

### A. The email dataset

Email dataset was created by combining a week's worth of emails (90 emails) from a user's sent box with outgoing emails (42 emails) sent by a computer infected with the *netsky-d* worm. The aim of this experiment is to correctly detect the 42 worm infected emails. With expert experience, we decide to use these attributes of each individual email: its sender is spoofed or not; whether it contains dangerous attachments, whether it contains non-dangerous attachments, the time interval since last email was sent. This information is listed in Table 3. Spoofed sender can be defined as a sender with a fake email address; pif: program information file, i.e. a dangerous file type; doc: word file (Considered as non-dangerous here).

TABLE 3 SUMMARY OF THE EMAIL DATASET

| Message | Spoofed Sender | number & type of attachments |
|---|---|---|
| 39-59 (worm) | Yes | 1 pif |
| 61-81 (worm) | Yes | 1 pif |
| 12, 101 | No | 1 doc |
| Others | No | 0 |

### B. The worm detection process

The four features used here are: signal1 – the time interval from last message sent, signal2 – the sender (spoofed-1, normal-0), signal3 – whether there are any dangerous attachment files (0 – no, 1 – yes), signal4 – whether there are some non-dangerous attachment files (0 - no, 1 - yes). Signal1 has a big range and varies from 0 to 94665, its *bpa* function is presented in Figure 8, the *bpa* functions used by the other three features are described in Table5.

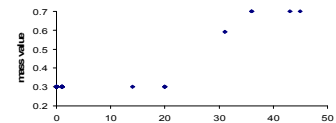

Fig. 8. *bpa* for signal1, the x-axis shows seconds since last email

With experience, the signals can be ranked in order of importance (high to low): signal2, signal3, signal1 and signal4. Threshold settings were decided following this order (shown in Table 4). From this information, mass value settings were derived, as shown in Table 5.

TABLE 4 THRESHOLD SETTINGS FOR EMAIL DATASET

| Features | threshold | Min m(normal) | Max m(normal) |
|---|---|---|---|
| Signal 1 | 30 | 0.3 | 0.7 |
| Signal 2 | 1 | 0.1 | 0.9 |
| Signal 3 | 1 | 0.2 | 0.8 |
| Signal 4 | 1 | 0.4 | 0.6 |

TABLE 5 MASS VALUE SETTINGS FOR EMAIL DATASET ( M($\Theta$)=0.01)

| | value | m(normal) | m(abnormal) |
|---|---|---|---|
| Signal 1 | =value | $0.4(1+e^{(value-30)})^{-1}+0.3$ | 1-m(normal)- m($\Theta$) |
| Signal 2 | =0 | 0.9 | 0.09 |
| | =1 | 0.1 | 0.89 |
| Signal 3 | =0 | 0.8 | 0.19 |
| | =1 | 0.2 | 0.79 |
| Signal 4 | =0 | 0.6 | 0.39 |
| | =1 | 0.4 | 0.59 |

We assign the mass values for each individual feature of one email using the settings in Table 5. These mass values for the same email are combined using the combination rule based on each individual hypothesis. The overall mass values of each hypothesis are generated. That email is classified as the hypothesis with the higher overall mass value.

### C. Experimental Results with email dataset

When using four signals, all 42 worm infected emails were detected correctly, as shown in Figure 9. As it is not always easy to determine a spoofed sender, we re-run the experiments removing this signal (Figure 10). Messages 39 and 61 were undetected when using the three remaining features (i.e. signal1, signal3, and signal4). The wrongly classified messages are those messages sent directly after legitimate traffic. Hence, the time intervals since last message sent of the two messages appear normal, with the only abnormal features being the executable attachment. Because all the three features have similar weights, and two of them indicate that the emails are normal, they are wrongly classified as normal. This can be corrected by weighting these features with greater different mass values, or adding in more effective features. A more effective feature can be the number of words contained in an email. We will look into these issues in future experiments.

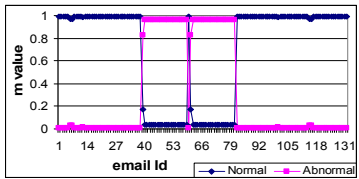

Fig. 9. Results of email data using all four features

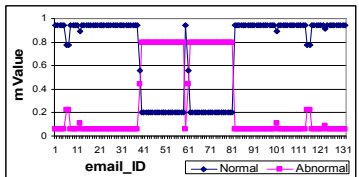

Fig. 10. Results of email data using signal1, signal3, signal4

### VII. CONCLUSIONS

The experimental results for the WBCD (with nine features and two classes) show that we successfully classify a standard dataset by combing multiple features using the D-S method. The results for the Iris dataset (with four features, three classes) show that we can also use D-S for problems with more than two classes, with fewer features. Our system successfully detects email worms through experiments with a realistic email dataset. These results indicate that D-S method works successfully for anomaly detection by combing the beliefs from multiple sources. Based on these results, we can conclude that D-S can be a promising method for network security problems with multiple features (from various data sources) and two or more classes.

Of course, like other classification algorithms, the initial feature selection influences overall performance. However, due to the inherent robustness of D-S, as long as there the majority features are suitable, our system still works, even if some features are poor. Furthermore, our approach works in situations where some feature values are missing, which is likely to occur in real world network security scenarios. Our continuing aim is to find out how D-S based algorithms can be used more effectively for the purpose of anomaly detection within the domain of network security.

### VIII. ACKNOWLEDGEMENTS

The authors would like to thank Julie Greensmith for useful comments and Jamie Twycross for collecting the email dataset.